\documentclass[10pt,twocolumn,letterpaper]{article}

\usepackage{cvpr}
\usepackage{times}
\usepackage{epsfig}
\usepackage{graphicx}
\usepackage{amsmath}
\usepackage{amssymb}
\usepackage{bbding}

\usepackage[breaklinks=true,bookmarks=false]{hyperref}

\cvprfinalcopy 

\def\cvprPaperID{****} 

\begin{document}

\title{Triply Supervised Decoder Networks for Joint Detection and Segmentation}

\author{Jiale Cao$^1$,~~Yanwei Pang$^{1*}$,~~Xuelong Li$^2$\\
$^1$ School of Electrical and Information Engineering, Tianjin University\\
$^2$Xi'an Institute of Optics and Precision Mechanics, Chinese Academy of Sciences\\
{\tt\small connor@tju.edu.cn, pyw@tju.edu.cn, xuelong\_li@opt.ac.cn}
}
\maketitle

\begin{abstract}
Joint object detection and semantic segmentation can be applied to many fields, such as self-driving cars and unmanned surface vessels. An initial and important progress towards this goal has been achieved by simply sharing the deep convolutional features for the two tasks. However, this simple scheme is unable to make full use of the fact that detection and segmentation are mutually beneficial. To overcome this drawback, we propose a framework called TripleNet where triple supervisions including detection-oriented supervision, class-aware segmentation supervision, and class-agnostic segmentation supervision are imposed on each layer of the decoder network. Class-agnostic segmentation supervision provides an objectness prior knowledge for both semantic segmentation and object detection.  Besides the three types of supervisions, two light-weight modules (i.e., inner-connected module and attention skip-layer fusion) are also incorporated into each layer of the decoder. In the proposed framework, detection and segmentation can sufficiently boost each other. Moreover, class-agnostic and class-aware segmentation on each decoder layer are not performed at the test stage. Therefore, no extra computational costs are introduced at the test stage. Experimental results on the VOC2007 and VOC2012 datasets demonstrate that the proposed TripleNet is able to improve both the detection and segmentation accuracies without adding extra computational costs. 
\end{abstract}

\section{Introduction}
\label{secIntroduction}
Object detection and semantic segmentation are two fundamental and important tasks in the field of computer vision. In recent few years, object detection \cite{Ren_FasterRCNN_NIPS_2015,Liu_SSD_ECCV_2016,Lin_Focal_ICCV_2017} and semantic segmentation \cite{Long_FCN_CVPR_2015,Chen_Deeplab_PAMI_2017,Badrinarayanan_SegNet_PAMI_2017} with deep convolutional networks \cite{Krizhevsky_ImageNet_NIPS_2012,Simonyan_VGG_arXiv_2014,He_MaskRCNN_ICCV_2017,Huang_DenseNet_CVPR_2017} have achieved great progress, respectively. Most state-of-the-art methods only focus on one single task, which does not join object detection and semantic segmentation together. However, joint object detection and semantic segmentation is very necessary and important in many applications, such as self-driving cars and unmanned surface vessels.

In fact, object detection and semantic segmentation are highly related. On the one hand, semantic segmentation usually used as a multi-task supervision can help improve object detection \cite{Mao_Hyper_CVPR_2017,Lin_Graininess_ECCV_2018}. On the other hand, object detection can be used as a prior knowledge to help improve performance of semantic segmentation \cite{He_MaskRCNN_ICCV_2017,Pinheiro_ROS_ECCV_2016}. 

\begin{figure}
\begin{center}
\includegraphics[width=3.3in]{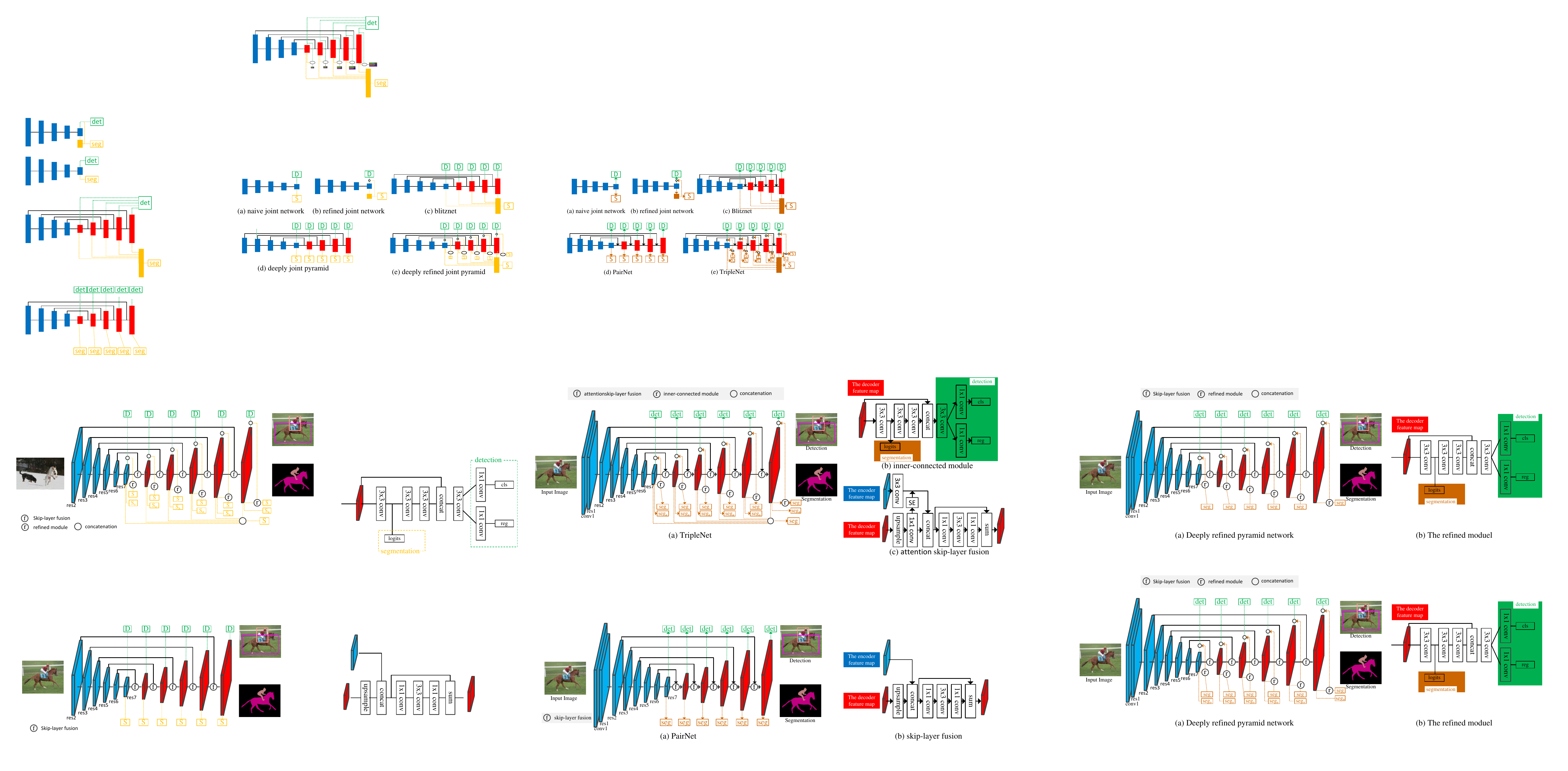}
\end{center}
\caption{Some architectures of joint detection and segmentation. (a) The last layer of the encoder is used for detection and segmentation\cite{Brazil_SDSRCNN_ICCV_2017}. (b) The branch for detection is refined by the branch for segmentation \cite{Mao_Hyper_CVPR_2017,Zhang_DES_CVPR_2018}. (c) Each layer of the decoder detects objects of different scales, and the fused layer is for segmentation \cite{Dvornik_Blitznet_ICCV_2017}. (d) The proposed PairNet. Each layer of the decoder is simultaneously for detection and segmentation. (e) The proposed TripleNet, which has three types of supervisions and some light-weight modules.}
\label{fig01}
\end{figure}

Due to application requirements and task relevance, joint object detection and semantic segmentation has gradually attracted the attention of researchers. Fig. \ref{fig01} summarizes three typical methods of joint object detection and semantic segmentation. Fig. \ref{fig01}(a) shows the simplest and most naive way where one branch for object detection and one branch for semantic segmentation are in parallel attached to the last layers of the encoder \cite{Brazil_SDSRCNN_ICCV_2017}.  In Fig. \ref{fig01}(b), the branch for object detection is further refined by the features from the branch for semantic segmentation \cite{Mao_Hyper_CVPR_2017,Zhang_DES_CVPR_2018}. Recently, the encoder-decoder network is further used for joint object detection and semantic segmentation. In Fig. \ref{fig01}(c), each layer of the decoder is used for multi-scale object detection, and the concatenated feature map from different layers of the decoder is used for semantic segmentation \cite{Dvornik_Blitznet_ICCV_2017}. The above methods have achieved great success for detection and segmentation. However, the performance is still far from the strict demand of real applications such as self-driving cars and unmanned surface vessels. One possible reason is that the mutual benefit between the two tasks is not fully exploited.

To exploit mutual benefit for joint object detection and semantic segmentation tasks, in this paper, we propose to impose three types of supervisions (i.e., detection-oriented supervision, class-aware segmentation supervision, and class-agnostic segmentation supervision) on each layer of the decoder network. Meanwhile, the light-weight modules (i.e., the inner-connected module and attention skip-layer fusion) are also incorporated. The corresponding network is called TripleNet (see Fig. \ref{fig01}(e)). It is noted that we also propose to only impose the detection-oriented supervision and class-aware segmentation supervision on each layer of the decoder, which is called PairNet (see Fig. \ref{fig01}(d)). The contributions of this paper can be summarized as follows:

(1) Two novel frameworks (i.e., PairNet and TripleNet) for joint object detection and semantic segmentation are proposed. In TripleNet, the detection-oriented supervision, class-aware segmentation supervision, and class-agnostic segmentation supervision are imposed on each layer of the decoder. Meanwhile, two light-weight modules (i.e., the inner-connected module and attention skip-layer fusion) is also incorporated into each layer of the decoder.

(2) A lot of synergies are gained from TripleNet. Both detection and segmentation accuracies are significantly improved. The improvement is not at expense of incurring extra computational costs because the class-agnostic segmentation and class-aware segmentation are not performed in each layer of the decoder at the test stage .

(3)	Experiments on the VOC 2007 and VOC 2012 datasets are conducted to demonstrate the effectiveness of the proposed TripleNet. 

The rest of this paper is organized as follows. Section \ref{secRelatedWork} reviews some related works of object detection and semantic segmentation. Section \ref{secOurMethods} introduces our proposed method in detail. Experiments are shown in Section \ref{secExperiments}. Finally, it is concluded in Section \ref{secConclusion}.

\section{Related works}
\label{secRelatedWork}
In this section, a review of object detection and semantic segmentation is firstly given. After that, some related works of joint object detection and semantic segmentation are further introduced.

\begin{figure*}[ht]
\begin{center}
\includegraphics{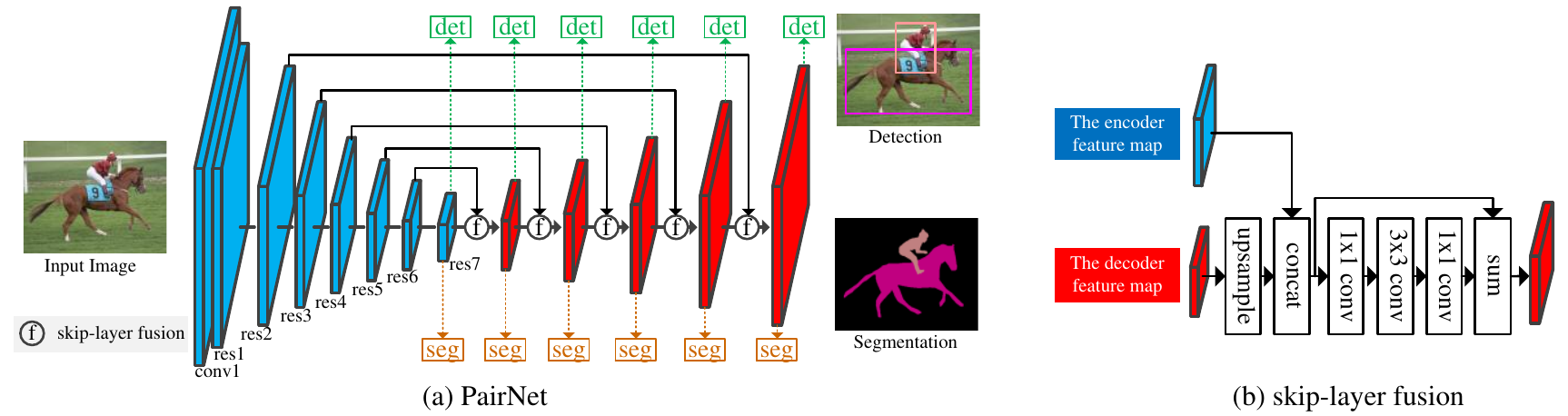}
\end{center}
\caption{The proposed PairNet for joint object detection and semantic segmentation. (a) The detailed architecture of PairNet. Each layer of the decoder is simultaneously used for detection and segmentation. (b) The skip-layer fusion used in PairNet.}
\label{fig02}
\end{figure*}
\textbf{Object detection} It aims to classify and locate objects in an image. Generally, the methods of object detection can be divided into two main classes: two-stage methods and one-stage methods. Two-stage methods firstly extract some candidate object proposals from an image and then classify these candidate proposals into the specific object categories. R-CNN \cite{Girshick_RCNN_CVPR_2014} and its variants (e.g., Fast RCNN \cite{Girshick_FastRCNN_ICCV_2015} and Faster RCNN \cite{Ren_FasterRCNN_NIPS_2015}) are the most representative frameworks among the two-stage methods. Based on R-CNN series, researchers have done many improvements \cite{Dai_RFCN_NIPS_2016,Lin_FPN_CVPR_2017,Cai_Cascade_CVPR_2018}. To accelerate detection speed, Dai \textit{et al.} \cite{Dai_RFCN_NIPS_2016} proposed R-FCN which uses position-sensitive feature maps for proposal classification and bounding box regression. To output multi-scale feature maps with strong semantics, Lin \textit{et al.} \cite{Lin_FPN_CVPR_2017} proposed feature pyramid network (FPN) based on skip-layer connection and top-down pathway. Recently, Cai \textit{et al.} \cite{Cai_Cascade_CVPR_2018} trained a sequence of object detectors with increasing IoU thresholds to improve detection quality.

One-stage methods directly predict object class and bounding box in a single network.  YOLO \cite{Redmon_YOLO_CVPR_2016} and SSD \cite{Liu_SSD_ECCV_2016} are two of the earliest proposed one-stage methods. After that, many variants are proposed \cite{Fu_DSSD_arXiv_2017,Kong_RON_CVPR_2017,Shen_DSOD_ICCV_2017,Zhou_STDN_CVPR_2018}. DSSD \cite{Fu_DSSD_arXiv_2017} and RON \cite{Kong_RON_CVPR_2017} use the encoder-decoder network to add context information for multi-scale object detection. To train object detector from scratch, DSOD \cite{Shen_DSOD_ICCV_2017} uses dense layer-wise connections on SSD for deep supervision. Instead of using in-network feature maps of different resolutions for multi-scale object detection, STDN \cite{Zhou_STDN_CVPR_2018} uses scale-transferrable module to generate different high-resolution feature maps from last feature map. To solve class imbalance in the training stage, RetinaNet \cite{Lin_Focal_ICCV_2017} introduces focal loss to downweight the contribution of easy samples.

\textbf{Semantic segmentation} It aims to predict the semantic label of each pixel in an image, which has achieved significant progress based on fully convolutional networks (i.e., FCN \cite{Long_FCN_CVPR_2015}). Generally, the methods of semantic segmentation can be also divided into two main classes: encoder-decoder methods and spatial pyramid methods. Encoder-decoder methods contain two subnetworks: an encoder subnetwork and a decoder subnetwork. The encoder subnetwork extracts strong semantic features and reduces spatial resolution of feature maps, which is usually based on the classical CNN models (e.g., VGG \cite{Simonyan_VGG_arXiv_2014}, ResNet \cite{He_ResNet_CVPR_2016}, DenseNet \cite{Huang_DenseNet_CVPR_2017}) pre-trained on ImageNet \cite{Russakovsky_ImageNet_IJCV_2015}. The decoder subnetwork gradually upsamples the feature maps of encoder subnetwork. DeconvNet \cite{Noh_DeconvNet_ICCV_2015} and SegNet \cite{Badrinarayanan_SegNet_PAMI_2017} use max-pooling indices of the encoder subnetwork to upsample the feature maps. To extract context information, some methods \cite{Peng_Largekernl_CVPR_2017,Lin_Refinenet_CVPR_2017,Yu_DFN_CVPR_2018} adopt skip-layer connection to combine the feature maps from the encoder and decoder subnetworks.

Spatial pyramid methods adopt the idea of spatial pyramid pooling \cite{He_SPP_ECCV_2014} to extract multi-scale information from the last output feature maps. Chen \textit{et al.} \cite{Chen_Deeplab_PAMI_2017,Chen_Deeplabv3_arXiv_2017,Wang_DUC_WACV_2017,Yang_DenseASPP_CVPR_2018} proposed to use multiple convolutional layers of different atrous rates in parallel (called ASPP) to extract multi-scale features. Instead of using convolutional layers of different atrous rates, Zhao \textit{et al.} \cite{Zhao_PSPNet_CVPR_2017} proposed pyramid pooling module (called PSPnet), which downsamples and upsamples the feature maps in parallel. Yang \textit{et al.} \cite{Yang_DenseASPP_CVPR_2018} proposed to use dense connection to cover object scale range densely.

\textbf{Joint object detection and semantic segmentation} It aims to simultaneously detect objects and predict pixel semantic labels by a single network. Recently, researchers have done some attempts. Yao \textit{et al.} \cite{Yao_JOSS_CVPR_2012} proposed to use the graphical model to holistic scene understanding. Teichmann \textit{et al.} \cite{Teichmann_Multinet_arXiv_2016} proposed to join object detection and semantic segmentation by sharing the encoder subnetwork. Kokkinos \cite{Kokkions_UberNet_CVPR_2017} also proposed to integrate multiple computer vision tasks together. Mao \textit{et al.} \cite{Mao_Hyper_CVPR_2017} found that joint semantic segmentation and pedestrian detection can help improve performance of pedestrian detection. The similar conclusion is also demonstrated by SDS-RCNN \cite{Brazil_SDSRCNN_ICCV_2017}. Meanwhile, joint instance semantic segmentation and object detection is also proposed \cite{Dvornik_Blitznet_ICCV_2017}.
Recently, Dvornik \textit{et al.} \cite{Dvornik_Blitznet_ICCV_2017} proposed a real-time framework (called BlitzNet) for joint object detection and semantic segmentation. It is based on the encoder-decoder network, where each layer of the decoder is used to detect objects of different scales and multi-scale fused layer is used for semantic segmentation. 

Though joint object detection and semantic segmentation has been explored recently, we argue that there is still room for further improvement. Thus, in this paper, we give some more exploration on joint object detection and semantic segmentation.

\section{The proposed methods}
\label{secOurMethods}
In recent years, the fully convolutional networks (FCN) with encoder-decoder structure have achieved great success on object detection \cite{Lin_Focal_ICCV_2017,Fu_DSSD_arXiv_2017} and semantic segmentation \cite{Badrinarayanan_SegNet_PAMI_2017}, respectively. For example, DSSD \cite{Fu_DSSD_arXiv_2017,Pinheiro_ROS_ECCV_2016} and RetinaNet \cite{Lin_Focal_ICCV_2017} use different layers of the decoder to detect objects of different scales, respectively. By using the  encoder-decoder structure, SegNet \cite{Badrinarayanan_SegNet_PAMI_2017} and LargeKernel \cite{Peng_Largekernl_CVPR_2017} generate high-resolution logits for semantic segmentation. Based on above observations, a very natural and simple idea is that FCN with encoder-decoder is suitable for joint object detection and semantic segmentation.

In this section, we give a detailed introduction about the proposed paired supervision decoder network (i.e., PairNet) and triply supervised decoder network (i.e., TripleNet) for joint object detection and semantic segmentation.

\begin{figure*}[t]
\begin{center}
\includegraphics{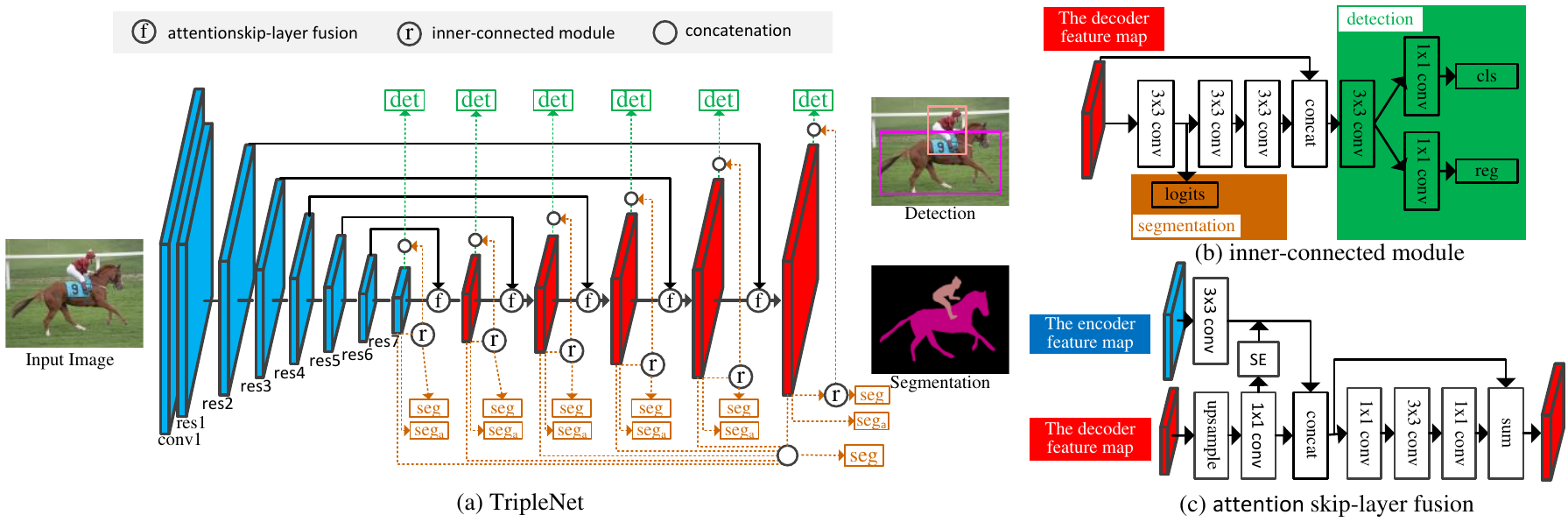}
\end{center}
\caption{The proposed TripleNet for joint object detection and semantic segmentation. (a) The detailed architecture of TripleNet. (b) The inner-connected module. (c) attention skip-layer fusion.}
\label{fig03}
\end{figure*}

\subsection{Paired supervision decoder network (PairNet)}
\label{secPairNet}
Based on the encoder-decoder structure, a feature pyramid network is naturally proposed to join object detection and semantic segmentation. Namely, the supervision of object detection and semantic segmentation is added to each layer of the decoder, which is called PairNet. On the one hand, PairNet uses different layers of the decoder to detect objects of different scales. On the other hand, instead of using the last high-resolution layer for semantic segmentation which is adopted by most state-of-the-art methods \cite{Badrinarayanan_SegNet_PAMI_2017,Peng_Largekernl_CVPR_2017}, PairNet uses each layer of the decoder to respectively parse pixel semantic labels.  Though the proposed PairNet is very simple and naive, it has not been explored for joint object detection and semantic segmentation to the best of our knowledge.

Fig. \ref{fig02}(a) gives the detailed architecture of PairNet. The input image firstly goes through a fully convolutional network with encoder-decoder structure. The encoder gradually down-samples the feature map. In this paper, the famous ResNet-50 or ResNet101 \cite{He_ResNet_CVPR_2016} (i.e., res1-res4) and some new added residual blocks (i.e., res5-res7) construct the encoder. The decoder gradually maps the low-resolution feature map to the high-resolution feature map. To enhance context information, skip-layer fusion is used to fuse the feature map from the decoder and the corresponding feature map from the encoder. Fig. \ref{fig02}(b) gives the illustration of skip-layer fusion. The feature maps in the decoder is firstly upsampled by bilinear interpolation and then concatenated with the corresponding feature maps of the same resolution in the encoder. After that, the concatenated feature maps go through a residual unit to generate the output feature maps.

To join object detection and semantic segmentation, each layer of the decoder is further split into two different branches. The branch of object detection consists of a $3\times 3$ convolutional layer and two sibling $1\times 1$ convolutional layers for object classification and bounding box regression. The branch of object detection at different layers is used to detect objects of different scales. Specifically, the branch at front layer of the decoder with low resolution is used to detect large-scale objects, while the branch at latter layer with high resolution is used to detect small-scale objects. 

The branch of semantic segmentation consists of a $3\times 3$ convolutional layer to generate the logits. There are two different ways to compute the segmentation loss. The first one is that the segmentation logits are upsampled to the same resolution of ground-truth, and the second one is that the ground-truth is downsampled to the same resolution of the logits. We found that the first strategy have a little better performance, which is adopted in the follows. 

\subsection{Triply supervised decoder network (TripleNet)}
To further improve the performance of joint object detection and semantic segmentation, triply supervision decoder network (called TripleNet) is further proposed, where detection-oriented supervision, class-aware segmentation supervision, and class-agnostic segmentation supervision are added on each layer of the decoder. Fig. \ref{fig03}(a) gives the detailed architecture of TripleNet.  Compared to PairNet, TripleNet add some new modules (i.e., multiscale fused segmentation, the inner-connected module, and class-agnostic segmentation supervision).  In the following section, we introduce these modules in detailed.

\textbf{Multiscale fused segmentation} It has been demonstrated that multi-scale features are useful for semantic segmentation \cite{Dvornik_Blitznet_ICCV_2017,Zhao_PSPNet_CVPR_2017,Yu_Dilate_ICLR_2016}. To use multi-scale features of different layers in the decoder for better semantic segmentation, the feature maps of different layers in the decoder are upsampled to the same spatial resolution and concatenated together. After that, a $3\times 3$ convolutional layer is used to generate the segmentation logits. Compared to the segmentations based on one layer of the decoder, multilayer fused features can make better use of context information. Thus, multilayer fused segmentation is used for final prediction at the test stage. Meanwhile, the semantic segmentation based on each layer of the decoder can be seen as a deep supervision for feature learning.

\textbf{The inner-connected module} In Section \ref{secPairNet}, PairNet only shares the base network for object detection and semantic segmentation, while the branches of object detection and semantic segmentation have no cross. To further help object detection, an inner-connected module is proposed to refine object detection by the logits of semantic segmentation.  Fig. \ref{fig03}(b) shows the inner-connected module in layer $i$. The feature map in layer $i$ first goes through a $3\times 3$ convolutional layer to produce the segmentation logits for the branch of semantic segmentation. Meanwhile, the segmentation logit goes through two $3\times 3$ convolutional layers to generate new feature map which are further concatenated with feature maps in layer $i$. Based on concatenated feature maps, a $3\times 3$ convolutional layer is used to generate the feature map for the branch of object detection. 

\begin{table*}
\renewcommand{\arraystretch}{1.2}
\footnotesize
\centering
\begin{center}
\setlength{\tabcolsep}{4mm}{
\begin{tabular}{l|ccc|cccc|c|c}
\hline
Method & det        & seg\_fine & seg\_all & MFS     & IC      & CAS    & ASF  & mAP     & mIoU   \\
\hline
(a) only detection & \Checkmark &           &          &         &         &        &  & 78.0    & N/A    \\
(b) only segmentation with fine layer &         & \Checkmark   &          &         &         &       &   & N/A     & 72.5   \\
(c) only segmentation with all layers &         &           & \Checkmark  &         &         &       &   & N/A     & 72.9   \\
(d) PairNet & \Checkmark &           & \Checkmark  &         &         &     &     & 78.9    & 73.1   \\
\hline
(e) add MFS & \Checkmark &           & \Checkmark  & \Checkmark &         &     &    & 79.0        & 73.5    \\
(f) add MFC and IC & \Checkmark &           & \Checkmark  & \Checkmark & \Checkmark &     &    & 79.5         & 73.6         \\
(g) add MFC, IC, and ASF & \Checkmark &           & \Checkmark  & \Checkmark & \Checkmark & \Checkmark & & 79.7        & 74.4        \\
(g) TripleNet & \Checkmark &           & \Checkmark  & \Checkmark & \Checkmark & \Checkmark & \Checkmark & 80.0        & 74.8        \\
\hline
\end{tabular}}
\end{center}
\caption{Ablation experiments of PairNet and TripleNet on the \texttt{VOC2012-val-seg} set. The backbone model is ResNet50 \cite{He_ResNet_CVPR_2016}, and the input image is rescaled to the size of $300\times 300$. ``MFS'' means multiscale fused segmentation, ``IC'' means inner-connected module, ``CAS'' means class-aware segmentation, and ``ASF'' means attention skip-layer fusion.}
\label{tab01}
\end{table*}
\textbf{Class-agnostic segmentation supervision} Semantic segmentation mentioned above is class-aware, which aims to simultaneously identify specific object categories and the background. We argue that class-aware semantic segmentation may ignore the discrimination between objects and the background. Therefore, class-agnostic segmentation supervision module is further added to each layer of the decoder. Specifically, a $3\times 3$ convolutional layer is added to generate the logits of class-agnostic semantic segmentation. To generate the ground-truth of class-agnostic semantic segmentation, the objects of different categories are set as one category, and the background is set as another category.

\textbf{Attention skip-layer fusion} In Section \ref{secPairNet}, PairNet simply fuses the feature maps of the decoder and the corresponding feature maps of the encoder. Generally, the features from the layer of the encoder have relatively low-level semantic, and that from the layer of decoder have relatively high-level semantic. To enhance informative features and suppress less useful features from the encoder by the features from the decoder, Squeeze-and-Excitation (SE) \cite{Hu_SENet_CVPR_2017} block is used. The input of a SE block is the layer of the decoder, and the output of SE block is used to scale the layer of the encoder. After that, the layer of the decoder and the scaled layer of the encoder is concatenated for fusion.     

\section{Experiments}
\label{secExperiments}
\subsection{Datasets}
To demonstrate the effectiveness of proposed methods and compare with same state-of-the-art methods, some experiments on the famous VOC 2007 and VOC 2012 datasets \cite{Everingham_VOC_IJCV_2010} are conducted in this section. 

The PASCAL VOC challenge \cite{Everingham_VOC_IJCV_2010} has been held annually since 2006, which consists of three principal challenges (i.e., image classification, object detection, and semantic segmentation). Among these annual challenges, the VOC 2007 and VOC 2012 datasets are usually used to evaluate the performance of object detection and semantic segmentation, which have 20 object categories. The VOC 2007 dataset contains 5011 \texttt{trainval} images and 4952 \texttt{test} images. The VOC 2012 dataset is split into three subsets (i.e., \texttt{train}, \texttt{val}, and \texttt{test}). The \texttt{train} set contains 5717 images for detection and 1464 images for semantic segmentation (called \texttt{VOC12-train-seg}). The \texttt{val} set contains 5823 images for detection and 1449 images for segmentation (called \texttt{VOC12-val-seg}). The test set contains 10991 images for detection and 1456 for segmentation. To enlarge the training set for semantic segmentation, the additional segmentation data provided by \cite{Harihar_SBD_ICCV_2011} is used, which contains 10582 training images (called \texttt{VOC12-trainaug-seg}).

For object detection, mean average precision (i.e., mAP) is used to performance evaluation. On the PASCAL VOC datasets, mAP is calculated under the IoU threshold of 0.5. For semantic segmentation, mean intersection over union (i.e., mIoU) is used for performance evaluation.

\subsection{Ablation experiments on the VOC 2012 dataset}
In this subsection, experiments are conducted on the PASCAL VOC 2012 to validate the effectiveness of proposed method.
On the PASCAL VOC 2012, the set of \texttt{VOC12-trainaug-seg} is used for training and the set of \texttt{VOC12-val-seg} is used for performance evaluation, where they have the ground truth of both object detection and semantic segmentation. The input images are rescaled to the size of $300\times 300$, and the size of mini-batch is 32. The total number of iteration in the training stage is 40$k$, where the learning rate of first 25$k$ iterations is 0.0001, that of following 10$k$ iterations is 0.00001, and that of last 5$k$ iterations is 0.000001.  
\begin{figure*}[t]
\begin{center}
   \includegraphics[width=1.0\linewidth]{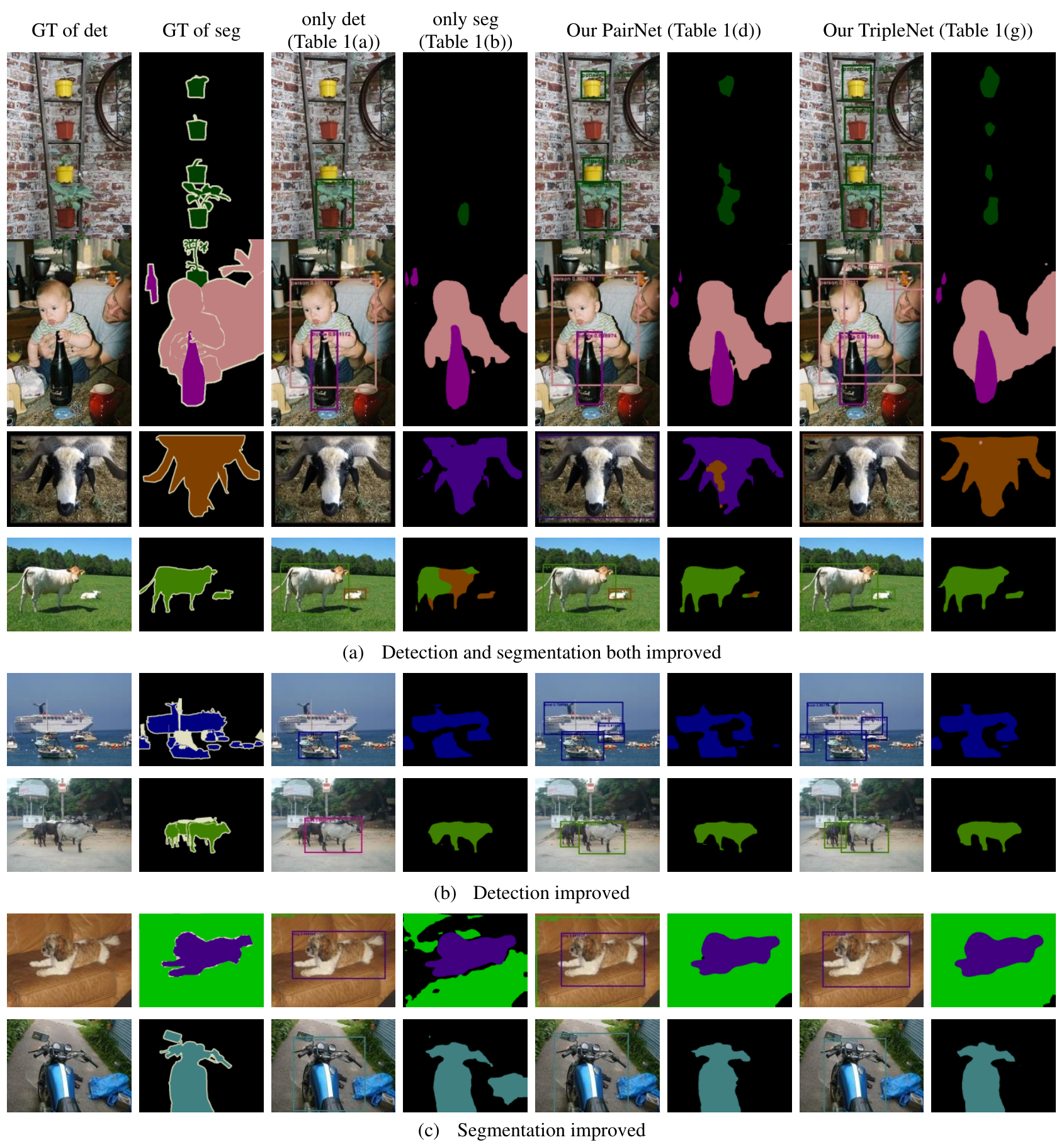}
\end{center}
   \caption{Visualization of detection or segmentation results of the methods in Table 1 (i.e., ``only det'', ``only seg'', PairNet, and TripleNet). (a) demonstrates that detection and segmentation can be both improved by PairNet and TripleNet. (b) demonstrates that detection is mainly improved by PairNet or TripleNet. (c) demonstrates that segmentation is mainly improved by PairNet or TripleNet.}
\label{fig04}
\end{figure*}

The top part of Table \ref{tab01} shows the ablation experiments of PairNet. When all different layers of the decoder are only used for multi-scale object detection (i.e., Table \ref{tab01}(a)), mAP of object detection is 78.0\%. When all different layers of the decoder are used for semantic segmentation (i.e., Table \ref{tab01}(c)), mIoU of semantic segmentation is 72.5\%. When all the different layer of the decoder is used for object detection and semantic segmentation together (i.e., Table \ref{tab01}(d)), mAP and mIoU of PairNet are 78.9\% and 73.1\%, respectively. Namely, PairNet can improve both object detection and semantic segmentation, which indicates that joint object detection and semantic segmentation on each layer of the decoder is useful. Meanwhile, the method using all the different layers of the decoder for segmentation (i.e., Table \ref{tab01}(c)) has better performance than the method only using the last layer of the decoder for segmentation (i.e., Table \ref{tab01}(d)). The reason may be explained by that all the layers of the decoder for semantic segmentation can give a much deeper supervision.

The bottom part of Table \ref{tab01} shows the ablation experiments of TripleNet. Based on PairNet, TripleNet adds three modules (i.e., MFS, IC, CAS, and AFS). When adding the MFS module, TripleNet outperforms PairNet by 0.1\% on object detection and 0.4\% on semantic segmentation, respectively.  When adding the MFS and IC modules, TripleNet outperforms PairNet by 0.6\% on object detection and 0.5\% on semantic segmentation. When adding all four modules, TripleNet has the best detection performance and segmentation performance.

Fig. \ref{fig04} further visualizes the experimental results of some methods in Table \ref{tab01}. The first two columns are ground-truth of detection and segmentation.  The results of only detection in Table \ref{tab01}(a) and only segmentation in Table \ref{tab01}(c) are shown in the third and forth columns. The results of PairNet in Table \ref{tab01}(d) and TripleNet in Table \ref{tab01}(g) are shown in fifth to eighth columns. In Fig. \ref{fig04}(a), the examples of detection and segmentation both improved by joint detection and segmentation are given. For example, in the first row, ``only det" and ``only seg" both miss three potted plant, while PairNet only misses one potted plant and TripleNet does not miss any potted plant. In Fig. \ref{fig04}(b), the examples of detection result improved are shown. For example, in the first row, ``only detect" can only detect one ship, PairNet can detect three ships, and TripleNet can detect four ships. In Fig. \ref{fig04}(c), the examples of segmentation results improved are shown. For example, in the second row, ``only seg" recognize blue bag as motorbike, but PairNet and TripleNet can recognize the blue bag as background.

\begin{table}
\renewcommand{\arraystretch}{1.2}
\footnotesize
\begin{center}
\setlength{\tabcolsep}{5mm}{
\begin{tabular}{l|c|cc}
\hline
method & input size & mAP & mIoU  \\
\hline
BlitzNet300 & $300\times 300$ & 78.8  & 72.9   \\
PairNet300      & $300\times 300$ & 78.9  & 73.1   \\
TripleNet300     & $300\times 300$ & 80.0  & 74.8   \\
\hline
\end{tabular}}
\end{center}
\caption{Comparison of BlitzNet, the proposed PairNet, and the proposed TripleNet. All the methods are re-implemented in the same parameter settings. }
\label{tab02}
\end{table}

Meanwhile, the proposed PairNet and TripleNet are also compared to the related BlitzNet \cite{Dvornik_Blitznet_ICCV_2017}. For fair comparison, BltizNet are re-implemented in the similar parameter settings as the proposed PairNet and TripleNet. PairNet which simply joins detection and segmentation in each layer of the decoder has been already comparable with BlitzNet. TripleNet outperforms BlitzNet on both object detection and semantic segmentation, which demonstrates that the proposed method can make full use of the mutual information to improve the two tasks. 

\begin{table}[t]
\renewcommand{\arraystretch}{1.2}
\footnotesize
\begin{center}
\setlength{\tabcolsep}{4.5mm}{
 \begin{tabular}{l|l|cc}
  \hline
  method        & backbone     &  mAP  & mIoU \\\hline
  SSD300  \cite{Liu_SSD_ECCV_2016}      & VGG          &  75.4   &  -        \\
  SSD512  \cite{Liu_SSD_ECCV_2016}      & VGG          &    79.4   &  -          \\
  RON384  \cite{Kong_RON_CVPR_2017}      & VGG16        &    75.4   &  -              \\
  DSSD513 \cite{Fu_DSSD_arXiv_2017}      & ResNet101     & 80.0  &    -                     \\     
  DES300  \cite{Zhang_DES_CVPR_2018}      & VGG16         & 77.1  &     -                     \\
  DES512  \cite{Zhang_DES_CVPR_2018}      & VGG16        & 80.3  &    -                      \\
  RefineDet512 \cite{Zhang_RefineDet_CVPR_2018}  & ResNet101    & 80.0  &    -                \\
  DFPR   \cite{Kong_DFPR_ECCV_2018}        & ResNet101     & 81.1  &                                     \\\hline
  FCN    \cite{Long_FCN_CVPR_2015}       & VGG16    &  -    &  62.2                    \\
  ParseNet  \cite{Liu_ParseNet_ICLR_2016}    & VGG16    &   -   &  69.8               \\
  Deeplab v2 \cite{Chen_Deeplab_PAMI_2017}   & VGG16                 & -  &  71.6                    \\
  DPN    \cite{Liu_DPN_ICCV_2015}       & VGG-like      &   -   &   74.1                   \\
  RefineNet  \cite{Lin_Refinenet_CVPR_2017}    & ResNet101   &- & 82.4\\
  PSPNet  \cite{Zhao_PSPNet_CVPR_2017}      & ResNet101      & -     &   82.6                \\
  DFN   \cite{Yu_DFN_CVPR_2018}       & ResNet101                     &  - & 82.7                           \\\hline  
  BlitzNet512 \cite{Dvornik_Blitznet_ICCV_2017} & ResNet50                   & 79.0  &    75.6              \\  
  TripleNet512 & ResNet101                    &  81.0 & 82.9                    \\\hline 
\end{tabular}} 
\end{center}
\caption{Results of object detection (mAP) and semantic segmentation (mIoU) on VOC 2012 \texttt{test} set.}
 \label{tabVOC2012}
\end{table}
\subsection{Comparison with state-of-the-art methods on the VOC2012 test dataset}
In this section, the proposed PairNet is compared with some state-of-the-art methods on the VOC 2007 dataset. Among these methods, SSD \cite{Liu_SSD_ECCV_2016}, RON \cite{Kong_RON_CVPR_2017}, DSSD \cite{Fu_DSSD_arXiv_2017}, DES \cite{Zhang_DES_CVPR_2018}, RefineDet \cite{Zhang_RefineDet_CVPR_2018}, and DFPR \cite{Kong_DFPR_ECCV_2018} are only used for object detection, ParseNet \cite{Liu_ParseNet_ICLR_2016}, Deeplab V2 \cite{Chen_Deeplab_PAMI_2017}, DPN \cite{Liu_DPN_ICCV_2015}, RefineNet \cite{Lin_Refinenet_CVPR_2017}, PSPNet \cite{Zhao_PSPNet_CVPR_2017}, DFPN \cite{Yu_DFN_CVPR_2018} are only used for semantic segmentation. Table \ref{tab02} shows object detect results (mAP) and semantic segmentation results (mIoU) of these methods on th VOC2012 \texttt{test} set. It can been seen most state-of-the-art methods can only output detection results (i.e., SSD, RON, DSSD, DES, RefineDet, and DFPR) or segmentation result (i.e., FCN, ParseNet, DeepLab, DPN, PSPNet, and DFPN). Only BlitzNet and our proposed TripleNet can simultaneously output the results of object detection and semantic segmentation. mAP and mIoU of BlitzNet are 79.0\% and 75.6\%, while mAP and mIoU of TripleNet are 81.0\% and 82.9\%. Thus, TripleNet outperforms BlitzNet by 2.0\% on object detection and 7.3\% on semantic segmentation. It can be also seen that TripleNet almost achieves state-of-the-art performance on both object detection and semantic segmentation.

\begin{table*}[t]
\renewcommand{\arraystretch}{1.2}
\scriptsize
\centering
\begin{center}
\setlength{\tabcolsep}{1.1mm}{
\begin{tabular}{l|c|c|cccccccccccccccccccc}
  \hline
  method             & backbone  & mAP       & aero      & bike      & bird      & boat      & bottle    & bus       & car       & cat       & chair     & cow       & table     & dog       & horse     & mbike     & persn     & plant     & sheep     & sofa      & train     & tv        \\ \hline
  SSD300 \cite{Liu_SSD_ECCV_2016}             & VGG16     & 77.5      & 79.5      & 83.9      & 76.0      & 69.6      & 50.5      & 87.0      & 85.7      & 88.1      & 60.3      & 81.5      & 77.0      & 86.1      & 87.5      & 84.0      & 79.4      & 52.3      & 77.9      & 79.5      & 87.6      & 76.8      \\
  SSD512  \cite{Liu_SSD_ECCV_2016}           & VGG16     & 79.5      & 84.8      & 85.1      & 81.5      & 73.0      & 57.8      & 87.8      & 88.3      & 87.4      & 63.5      & 85.4      & 73.2      & 86.2      & 86.7      & 83.9      & 82.5      & 55.6      & 81.7      & 79.0      & 86.6      & 80.0      \\ 
  DES300 \cite{Zhang_DES_CVPR_2018}      & VGG16     & 79.7      & 83.5      & 86.0      & 78.1      & 74.8      & 53.4      & 87.9      & 87.3      & 88.6      & 64.0      & 83.8      & 77.2      & 85.9      & 88.6      & 87.5 & 80.8      & 57.3      & 80.2      & 80.4      & 88.5 & 79.5      \\  
  DES512 \cite{Zhang_DES_CVPR_2018}      & VGG16     & 81.7 & 87.7 & 86.7      & 85.2 & 76.3 & 60.6      & 88.7 & 89.0 & 88.0      & 67.0 & 86.9      & 78.0 & 87.2      & 87.9      & 87.4      & 84.4 & 59.2 & 86.1 & 79.2      & 88.1      & 80.5      \\ 
  DSSD321 \cite{Fu_DSSD_arXiv_2017}      & ResNet101     & 78.6 & 81.9 & 84.9      & 80.5 & 68.4 & 53.9      & 85.6 & 86.2 & 88.9      & 61.1 & 83.5      & 78.7 & 86.7      & 88.7      & 86.7      & 79.7 & 51.7 & 78.0 & 80.9      & 87.2      & 79.4      \\   
  DSSD513 \cite{Fu_DSSD_arXiv_2017}      & ResNet101     & 81.5 & 86.6 & 86.2      & 82.6 & 74.9 & 62.5      & 89.0 & 88.7 & 88.8      & 65.2 & 87.0      & 78.7 & 88.2      & 89.0      & 87.5      & 83.7 & 51.1 & 86.3 & 81.6      & 85.7      & 83.7      \\ 
  STDN300 \cite{Zhou_STDN_CVPR_2018}    & DenseNet169     & 78.1 & 81.1 & 86.9      & 76.4 & 69.2 & 52.4      & 87.7 & 84.2 & 88.3      & 60.2 & 81.3      & 77.6 & 86.6      & 88.9      & 87.8      & 76.8 & 51.8 & 78.4 & 81.3      & 87.5      & 77.8      \\  
  STDN513 \cite{Zhou_STDN_CVPR_2018}    & DenseNet169     & 80.9 & 86.1 & 89.3      & 79.5 & 74.3 & 61.9      & 88.5 & 88.3 & 89.4      & 67.4 & 86.5      & 79.5 & 86.4      & 89.2      & 88.5      & 79.3 & 53.0 & 77.9 & 81.4      & 86.6      & 85.5      \\  
  BlitzNet300 \cite{Dvornik_Blitznet_ICCV_2017}    & ResNet50     & 79.1 & 86.7 & 86.2      & 78.9 & 73.1 & 47.6      & 85.7 & 86.1 & 87.7      & 59.3 & 85.1      & 78.4 & 86.3      & 87.9      & 84.2      & 79.1 & 58.5 & 82.5 & 81.7      & 85.7     & 81.8      \\  
  BlitzNet512 \cite{Dvornik_Blitznet_ICCV_2017}    & ResNet50     & 81.5 & 87.0 & 87.6      & 83.5 & 75.7 & 59.1      & 87.6 & 88.0 & 88.8      & 64.1& 88.4      & 80.9 & 87.5      & 88.5      & 86.9      & 81.5 & 60.6 & 86.5 & 79.3      & 87.5      & 81.7      \\
  RefineDet320 \cite{Zhang_RefineDet_CVPR_2018}    & VGG16     & 80.0 & - & -   & - & - & -    & - & - & -      & - & -  & - & -   & -    & -   & - & - & - & -      & -     & -      \\  
  RefineDet512 \cite{Zhang_RefineDet_CVPR_2018}    & VGG16     & 81.8 & - & -      & - & - & -      & - & - & -      & - & -      & - & -    & -      & -      & - & - & - & -      & -      & -      \\  
  DFPR300  \cite{Kong_DFPR_ECCV_2018}           & ResNet101 & 77.1      & 89.3      & 84.9      & 79.9      & 75.6      & 55.4      & 88.2      & 88.6      & 88.6      & 63.3      & 87.9      & 78.8      & 87.3      & 87.7      & 85.5      & 80.5      & 55.4      & 81.1      & 79.6 & 87.8      & 78.5      \\  
  DFPR512  \cite{Kong_DFPR_ECCV_2018}           & ResNet101 & 82.4      & 92.0      & 88.2 & 81.1      & 71.2      & 65.7      & 88.2      & 87.9      & 92.2      & 65.8      & 86.5      & 79.4      & 90.3      & 90.4      & 89.3 & 88.6      & 59.4      & 88.4      & 75.3      & 89.2      & 78.5 \\  
   \hline          
  TripleNet300~    & ResNet50     & 79.3 & 81.4 & 85.0      & 79.5 & 72.1 & 53.7      & 85.3 & 85.9 & 87.8      & 62.5 & 85.1      & 78.7 & 87.8      & 88.6      & 85.7      & 79.5 & 56.8 & 80.7 & 79.2      & 88.7      & 81.4      \\ 
  TripleNet512~    & ResNet50     & 82.4 & 88.9 & 86.9      & 85.0 & 77.5 & 61.5      & 87.7 & 88.2 & 89.2      & 66.0 & 88.3      & 79.6 & 87.5      & 88.8      & 87.3      & 82.3 & 62.4 & 86.1 & 81.7      & 89.2      & 82.4      \\   
  TripleNet512~    & ResNet101     & 82.7 & 88.9 & 87.8      & 83.7 & 79.6 & 62.9      & 87.9 & 88.3 & 88.5      & 67.5 & 89.1      & 81.2 & 88.0      & 89.5      & 87.9      & 83.3 & 58.7 & 85.1 & 83.4      & 88.8      & 84.1      \\ \hline  
\end{tabular}}
\end{center}
\caption{Results of object detection (mAP) on the VOC 2007 \texttt{test} set.}
\label{tab03}
\end{table*}
\subsection{Comparison with some state-of-the-art methods on the VOC 2007 test dataset}
In this section, the proposed TripleNet and some state-of-the-art methods (i.e., SSD \cite{Liu_SSD_ECCV_2016}, DES \cite{Zhang_DES_CVPR_2018}, DSSD \cite{Fu_DSSD_arXiv_2017}, STDN \cite{Zhou_STDN_CVPR_2018}, BlitzNet \cite{Dvornik_Blitznet_ICCV_2017}, RefineDet \cite{Lin_Refinenet_CVPR_2017}), and DFPR \cite{Kong_DFPR_ECCV_2018} are further compared on the VOC 2007 \texttt{test} set. Because only the ground-truth of object detection is provided, these methods are only evaluated on object detection. Table \ref{tab03} shows mAP of these methods. mAP of TripleNet is 82.7\%, which is higher than that of all state-of-the-art methods.

\section{Conclusion}
\label{secConclusion}
In this paper, we proposed two fully convolutional networks (i.e., PairNet and TripleNet) for joint object detection and semantic segmentation. PairNet simultaneously predicts objects of different scales by different layers and parses pixel semantic labels by all different layers. TripleNet adds four modules (i.e, multiscale fused segmentation, inner-connected module, class-agnostic segmentation supervision, and attention skip-layer fusion) to PairNet. Experiments demonstrate that TripleNet can achieve state-of-the-art performance on both object detection and semantic segmentation.


\begin{thebibliography}{10}\itemsep=-1pt

\bibitem{Badrinarayanan_SegNet_PAMI_2017}
V.~Badrinarayanan, A.~Kendall, and R.~Cipolla.
\newblock Segnet: A deep convolutional encoder-decoder architecture for image segmentation.
\newblock {\em IEEE Transactions on Pattern Analysis and Machine Intelligence},  39(12):2481--2495, 2017.

\bibitem{Brazil_SDSRCNN_ICCV_2017}  
G. Brazil, X. Yin, and X. Liu.
Illuminating pedestrians via simultaneous detection \& segmentation.
\newblock In {\em Proc. IEEE International Conference on Computer Vision}, 2017.

\bibitem{Chen_Deeplabv3_arXiv_2017}
L.-C. Chen, G.~Papandreou, F.~Schroff, and H.~Adam.
\newblock Rethinking atrous convolution for semantic image segmentation.
\newblock {\em arXiv:1706.05587}, 2017.

\bibitem{Cai_Cascade_CVPR_2018} 
Z. Cai and N. Vasconcelos.
Cascade R-CNN: Delving into high quality object detection.
\newblock In {\em Proc. IEEE Conference on Computer Vision and Pattern Recognition},  2018.

\bibitem{Chen_Deeplab_PAMI_2017}
L.-C. Chen, G.~Papandreou, I.~Kokkinos, K.~Murphy, and A.~L. Yuille.
\newblock Deeplab: Semantic image segmentation with deep convolutional nets, atrous convolution, and fully connected crfs.
\newblock {\em IEEE Transactions on Pattern Analysis and Machine Intelligence}, 40(4):834-848, 2017.

\bibitem{Dai_RFCN_NIPS_2016}
J.~Dai, Y.~Li, K.~He, and J.~Sun.
\newblock R-fcn: Object detection via region-based fully convolutional  networks.
\newblock In {\em Proc. Advances in Neural Information Processing Systems}, 2016.

\bibitem{Dvornik_Blitznet_ICCV_2017} 
N. Dvornik, K. Shmelkov, J. Mairal, and C. Schmid.
BlitzNet: A real-time deep network for scene understanding.
\newblock In {\em Proc. IEEE International Conference on Computer Vision}, 2017.

\bibitem{Everingham_VOC_IJCV_2010}
M.~Everingham, L.~Van~Gool, C.~K. Williams, J.~Winn, and A.~Zisserman.
\newblock The pascal visual object classes (voc) challenge.
\newblock {\em International Journal of Computer Vision}, 88(2):303--338, 2010.

\bibitem{Fu_DSSD_arXiv_2017}
C.-Y. Fu, W.~Liu, A.~Ranga, A.~Tyagi, and A.~C. Berg.
\newblock Dssd: Deconvolutional single shot detector.
\newblock {\em arXiv:1701.06659}, 2017.

\bibitem{Girshick_FastRCNN_ICCV_2015}
R.~Girshick.
\newblock Fast r-cnn.
\newblock In {\em Proc. IEEE International Conference on Computer Vision}, 2015.

\bibitem{Girshick_RCNN_CVPR_2014}
R.~Girshick, J.~Donahue, T.~Darrell, and J.~Malik.
\newblock Rich feature hierarchies for accurate object detection and semantic segmentation.
\newblock In {\em Proc. IEEE Conference on Computer Vision and  Pattern Recognition}, 2014.

\bibitem{Harihar_SBD_ICCV_2011}
B.~Hariharan, P.~Arbel{\'a}ez, L.~Bourdev, S.~Maji, and J.~Malik.
\newblock Semantic contours from inverse detectors.
\newblock In {\em Proc. IEEE International Conference on Computer Vision},  2011.
  
\bibitem{He_SPP_ECCV_2014}
K.~He, X.~Zhang, S.~Ren, and J.~Sun.
\newblock Spatial pyramid pooling in deep convolutional networks for visual recognition.
\newblock In {\em Proc. European Conference on Computer Vision}, 2014.  

\bibitem{He_MaskRCNN_ICCV_2017}
K.~He, G.~Gkioxari, P.~Doll{\'a}r, and R.~Girshick.
\newblock Mask r-cnn.
\newblock In {\em Proc. IEEE International Conference on Computer  Vision}, 2017.

\bibitem{He_ResNet_CVPR_2016}
K.~He, X.~Zhang, S.~Ren, and J.~Sun.
\newblock Deep residual learning for image recognition.
\newblock In {\em Proc. IEEE Conference on Computer Vision and Pattern Recognition}, 2016.

\bibitem{Hu_SENet_CVPR_2017}
J.~Hu, L.~Shen, and G.~Sun.
\newblock Squeeze-and-excitation networks.
\newblock In {\em Proc. IEEE Conference on Computer Vision and Pattern Recognition}, 2017.

\bibitem{Huang_DenseNet_CVPR_2017}
G.~Huang, Z.~Liu, L.~van~der Maaten, and K.~Q. Weinberger.
\newblock Densely connected convolutional networks.
\newblock In {\em Proc. IEEE Conference on Computer Vision and Pattern Recognition}, 2017.

\bibitem{Kong_RON_CVPR_2017}
T.~Kong, F.~Sun, A.~Yao, H.~Liu, M.~Lu, and Y.~Chen.
\newblock Ron: Reverse connection with objectness prior networks for object detection.
\newblock In {\em Proc. IEEE Conference on Computer Vision and Pattern Recognition}, 2017.

\bibitem{Kong_DFPR_ECCV_2018}
T. Kong, F. Sun, W. Huang, and H. Liu.
Deep feature pyramid reconguration for object detection.
\newblock In {\em Proc. European Conference on Computer Vision}, 2018. 

\bibitem{Krizhevsky_ImageNet_NIPS_2012}
A.~Krizhevsky, I.~Sutskever, and G.~E. Hinton.
\newblock Imagenet classification with deep convolutional neural networks.
\newblock In {\em Proc. Advances in Neural Information Processing Systems}, 2012.

\bibitem{Kokkions_UberNet_CVPR_2017}
I. Kokkinos.
\newblock UberNet: Training a universal convolutional neural network for low-, mid-, and high-level vision using
diverse datasets and limited memory.
\newblock In {\em Proc. IEEE Conference on Computer Vision and Pattern Recognition}, 2017.

\bibitem{Lin_FPN_CVPR_2017}
T.-Y. Lin, P. Doll´ar, R. Girshick, K. He, B. Hariharan, and S. Belongie.
\newblock Feature pyramid networks for object detection.
\newblock In {\em Proc. IEEE Conference on Computer Vision and Pattern Recognition}, 2017.  

\bibitem{Lin_COCO_ECCV_2014}
T.-Y. Lin, M.~Maire, S.~Belongie, J.~Hays, P.~Perona, D.~Ramanan,
  P.~Doll{\'a}r, and C.~L. Zitnick.
\newblock Microsoft coco: Common objects in context.
\newblock In {\em Proc. European Conference on Computer Vision}, 2014.

\bibitem{Lin_Graininess_ECCV_2018}  
C. Lin, J. Lu, G. Wang, and J. Zhou.
Graininess-aware deep feature learning for pedestrian detection.
\newblock In {\em Proc. European Conference on Computer Vision}, 2018. 

\bibitem{Lin_Refinenet_CVPR_2017}
G.~Lin, A.~Milan, C.~Shen, and I.~Reid.
\newblock Refinenet: Multi-path refinement networks with identity mappings for high-resolution semantic segmentation.
\newblock In {\em Proc. IEEE Conference on Computer Vision and Pattern Recognition}, 2017.

\bibitem{Lin_Focal_ICCV_2017}
T.-Y. Lin, P.~Goyal, R.~Girshick, K.~He, and P.~Doll{\'a}r.
\newblock Focal loss for dense object detection.
\newblock In {\em Proc. IEEE International Conference on Computer Vision}, 2017.

\bibitem{Liu_ParseNet_ICLR_2016}
W. Liu, A. Rabinovich, and A. C. Berg.
ParseNet: Looking wider to see better.
\newblock In {\em Proc. International Conference on Learning Representations}, 2016.

\bibitem{Liu_DPN_ICCV_2015}
Z. Liu, X. Li, P. Luo, C. Change Loy, and X. Tang.
Semantic image segmentation via deep parsing network.
\newblock In {\em Proc. IEEE International Conference on Computer Vision}, 2015.

\bibitem{Liu_SSD_ECCV_2016}
W.~Liu, D.~Anguelov, D.~Erhan, C.~Szegedy, S.~Reed, C.-Y. Fu, and A.~C. Berg.
\newblock Ssd: Single shot multibox detector.
\newblock In {\em Proc. European Conference on Computer Vision}, 2016.

\bibitem{Long_FCN_CVPR_2015}
J.~Long, E.~Shelhamer, and T.~Darrell.
\newblock Fully convolutional networks for semantic segmentation.
\newblock In {\em Proc. IEEE Conference on Computer Vision and  Pattern Recognition}, 2015.

\bibitem{Mao_Hyper_CVPR_2017}  
J. Mao, T. Xiao, Y. Jiang, and Z. Cao.
What can help pedestrian detection?
\newblock In {\em Proc. IEEE Conference on Computer Vision and Pattern Recognition},  2017.

\bibitem{Noh_DeconvNet_ICCV_2015} 
H. Noh, S. Hong, and B Han.
Learning deconvolution network for semantic segmentation.
\newblock In {\em Proc. IEEE International Conference on Computer Vision}, 2017.

\bibitem{Peng_Largekernl_CVPR_2017}
C.~Peng, X.~Zhang, G.~Yu, G.~Luo, and J.~Sun.
\newblock Large kernel matters--improve semantic segmentation by global convolutional network.
\newblock In {\em Proc. IEEE Conference on Computer Vision and Pattern Recognition},  2017.

\bibitem{Pinheiro_ROS_ECCV_2016} 
P. O. Pinheiro, T.-Y. Lin, R. Collobert, and P. Doll{\'a}r.
\newblock Learning to refine object segments.
\newblock In {\em Proc. European Conference on Computer Vision}, 2016.

\bibitem{Redmon_YOLO_CVPR_2016}
J.~Redmon, S.~Divvala, R.~Girshick, and A.~Farhadi.
\newblock You only look once: Unified, real-time object detection.
\newblock In {\em Proc. IEEE Conference on Computer Vision and Pattern Recognition}, 2016.

\bibitem{Ren_FasterRCNN_NIPS_2015}
S.~Ren, K.~He, R.~Girshick, and J.~Sun.
\newblock Faster r-cnn: Towards real-time object detection with region proposal networks.
\newblock In {\em Proc. Advances in Neural Information Processing Systems}, 2015.
  
\bibitem{Russakovsky_ImageNet_IJCV_2015}
O. Russakovsky, J. Deng, H. Su, J. Krause, S. Satheesh, S. Ma, Z. Huang, A. Karpathy, A. Khosla, M. Bernstein, A. C. Berg, and L. Fei-Fei.
ImageNet large scale visual recognition challenge.
\newblock {\em International Journal of Computer Vision}, 2015.

\bibitem{Shen_DSOD_ICCV_2017} 
Z. Shen, Z. Liu, J. Li, Y.-G. Jiang, Y. Chen, and X. Xue.
DSOD: Learning deeply supervised object detectors from scratch.
\newblock In {\em Proc. IEEE International Conference on Computer Vision}, 2017.

\bibitem{Simonyan_VGG_arXiv_2014}
K.~Simonyan and A.~Zisserman.
\newblock Very deep convolutional networks for large-scale image recognition.
\newblock {\em arXiv:1409.1556}, 2014.

\bibitem{Teichmann_Multinet_arXiv_2016} 
M. Teichmann, M. Weber, M. Zoellner, R. Cipolla, and R. Urtasun. 
\newblock Multinet: Real-time joint semantic reasoning
for autonomous driving. \newblock In {\em arXiv:1612.07695}, 2016.

\bibitem{Uijlings_SS_IJCV_2013}
J.~R. Uijlings, K.~E. Van De~Sande, T.~Gevers, and A.~W. Smeulders.
\newblock Selective search for object recognition.
\newblock {\em International journal of computer vision}, 104(2):154--171, 2013.

\bibitem{Wang_DUC_WACV_2017}
P.~Wang, P.~Chen, Y.~Yuan, D.~Liu, Z.~Huang, X.~Hou, and G.~Cottrell.
\newblock Understanding convolution for semantic segmentation.
\newblock In {\em Proc. IEEE Winter Conference on Applications of Computer Vision},  2017.

\bibitem{Yang_DenseASPP_CVPR_2018}
M. Yang, K. Yu, C. Zhang, Z. Li, and K. Yang.
DenseASPP for semantic segmentation in street scenes.
\newblock In {\em Proc. IEEE Conference on Computer Vision and Pattern Recognition},  2018.

\bibitem{Yao_JOSS_CVPR_2012} 
J. Yao, S. Fidler, and R. Urtasun. 
\newblock Describing the scene as a whole: Joint object detection, scene classification and semantic segmentation. 
\newblock In {\em Proc. IEEE Conference on Computer Vision and Pattern Recognition}, 2012.

\bibitem{Yu_DFN_CVPR_2018}
C. Yu, J. Wang, C. Peng, C. Gao, G. Yu, and N. Sang.
Learning a discriminative feature network for semantic segmentation.
\newblock In {\em Proc. IEEE Conference on Computer Vision and Pattern Recognition},  2018.

\bibitem{Yu_Dilate_ICLR_2016}
F.~Yu and V.~Koltun.
\newblock Multi-scale context aggregation by dilated convolutions.
\newblock In {\em Proc. International Conference on Learning Representations}, 2016.

\bibitem{Zhang_DES_CVPR_2018}
Z. Zhang, S. Qiao, C. Xie, W. Shen, B. Wang, and A. L. Yuille.
Single-shot object detection with enriched semantics.
\newblock In {\em Proc. IEEE Conference on Computer Vision and Pattern Recognition},  2018.

\bibitem{Zhang_RefineDet_CVPR_2018}
S. Zhang, L. Wen, X. Bian, Z. Lei, and S. Z. Li.
Single-shot refinement neural network for object detection.
\newblock In {\em Proc. IEEE Conference on Computer Vision and Pattern Recognition},  2018.
  
\bibitem{Zhao_PSPNet_CVPR_2017}
H.~Zhao, J.~Shi, X.~Qi, X.~Wang, and J.~Jia.
\newblock Pyramid scene parsing network.
\newblock In {\em Proc. IEEE Conference on Computer Vision and Pattern Recognition},  2017.

\bibitem{Zhou_STDN_CVPR_2018} 
P. Zhou, B. Ni, C. Geng, J. Hu, and Y. Xu.
Scale-transferrable object detection.
\newblock In {\em Proc. IEEE Conference on Computer Vision and Pattern Recognition},  2018.


\end{thebibliography}

\end{document}